\documentclass{article}

\usepackage[final]{neurips_2025}
\usepackage{amsmath}
\usepackage{multirow} 
\usepackage{graphicx} 
\usepackage{amssymb}

\usepackage[utf8]{inputenc} 
\usepackage[T1]{fontenc}    
\usepackage{hyperref}       

\usepackage[capitalize]{cleveref}
\crefname{section}{Sec.}{Secs.}
\Crefname{section}{Section}{Sections}
\crefname{table}{Tab.}{Tabs.}
\Crefname{table}{Table}{Tables}
\crefname{algocf}{Alg.}{Algs.}
\Crefname{algocf}{Algorithm}{Algorithms}

\usepackage{url}            
\usepackage{booktabs}       
\usepackage{amsfonts}       
\usepackage{nicefrac}       
\usepackage{microtype}      
\usepackage{xcolor}         
\usepackage{natbib}
\usepackage{bbm}

\title{Video-based Generalized Category Discovery via Memory-Guided Consistency-Aware Contrastive Learning}

\author{%
  Zhang Jing\textsuperscript{1,†} \\
  \texttt{zhangjing98@nudt.edu.cn} \\
  \And
  Pu Nan\textsuperscript{2,†} \\
  \texttt{nanpu@unitn.it} \\
  \And
  Xie Yuxiang\textsuperscript{1,*} \\
  \texttt{yxxie@nudt.edu.cn} \\
  \AND
  Guo Yanming\textsuperscript{1} \\
  \texttt{guoyanming@nudt.edu.cn} \\
  \And
  Lu Qianqi\textsuperscript{1} \\
  \texttt{luqianqi@nudt.edu.cn} \\
  \And
  Zou Shiwei\textsuperscript{1} \\
  \texttt{zsw0915@nudt.edu.cn} \\
  \And
  Yan Jie\textsuperscript{1} \\
  \texttt{yanjie@nudt.edu.cn} \\
  \And
  Chen Yan\textsuperscript{1} \\
  \texttt{chenyan0702@nudt.edu.cn} \\
  \vspace{0.5em}
  \small
  \textsuperscript{1}Laboratory for Big Data and Decision, National University of Defense Technology, Changsha, China \\
  \textsuperscript{2}Department of Information Engineering and Computer Science, University of Trento, Italy \\
  \vspace{0.2em}
  \textsuperscript{†}These authors contributed equally. \\
  \textsuperscript{*}Corresponding author: xieyuxiang@nudt.edu.cn
}
\begin{document}

\maketitle
\begin{abstract}
Generalized Category Discovery (GCD) is an emerging and challenging open-world problem that has garnered increasing attention in recent years. The goal of GCD is to categorize all samples in the unlabeled dataset, regardless of whether they belong to known classes or entirely novel ones. Most existing GCD methods focus on discovering categories in static images. However, relying solely on static visual content is often insufficient to reliably discover novel categories. For instance, bird species with highly similar appearances may exhibit distinctly different motion patterns. To bridge this gap, we extend the GCD problem to the video domain and introduce a new setting, termed \textbf{Video-GCD}. Compared with conventional GCD, which primarily focuses on how to leverage unlabeled image data, Video-GCD introduces additional challenges due to complex temporal and spatial dynamics. Thus,  effectively integrating multi-perspective information across time is crucial for accurate Video-GCD. To tackle this challenge, we propose a novel Memory-guided Consistency-aware Contrastive Learning (\textbf{MCCL}) framework, which explicitly captures temporal-spatial cues and incorporates them into contrastive learning through a consistency-guided voting mechanism. MCCL consists of two core components: Consistency-Aware Contrastive Learning(\textbf{CACL}) and Memory-Guided Representation Enhancement (\textbf{MGRE}). CACL exploits multi-perspective temporal features to estimate consistency scores between unlabeled instances, which are then used to weight the contrastive loss accordingly. MGRE introduces a dual-level memory buffer that maintains both feature-level and logit-level representations, providing global context to enhance intra-class compactness and inter-class separability. This in turn refines the consistency estimation in CACL, forming a mutually reinforcing feedback loop between representation learning and consistency modeling. To facilitate a comprehensive evaluation, we construct a new and challenging \textbf{Video-GCD benchmark}, which includes action recognition and bird classification video datasets. Extensive experiments demonstrate that our method significantly outperforms competitive GCD approaches adapted from image-based settings, highlighting the importance of temporal information for discovering novel categories in videos. The code will be publicly available.
\end{abstract}

\section{Introduction}

\begin{figure*}[!t]
\centering{
\includegraphics[width=13cm]{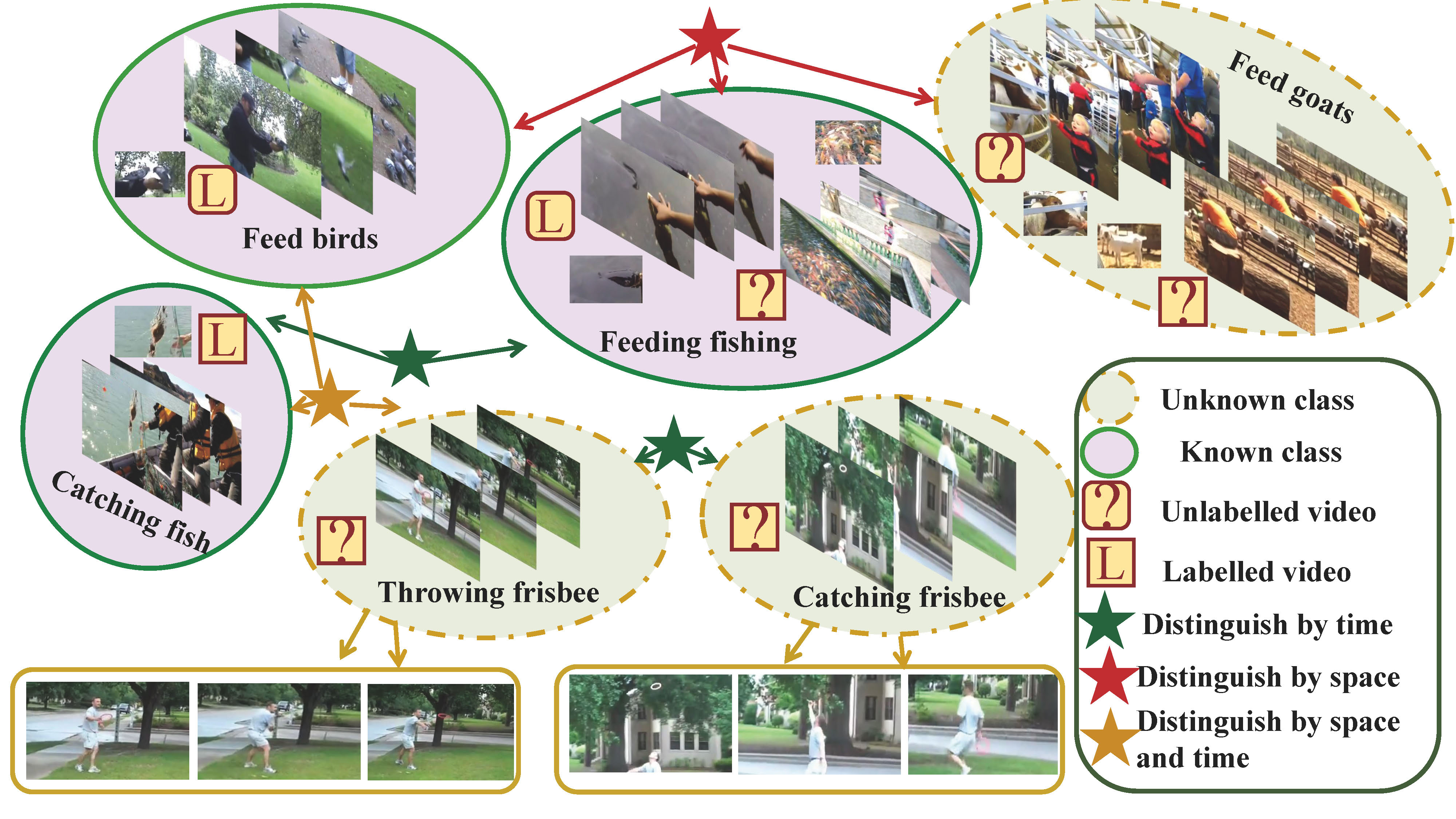}
}
\vspace{-0.5em}
\caption{\label{fig:1} Schematic diagram of the proposed video-GCD task.}
\vspace{-1.5em}
\end{figure*}

Generalized Category Discovery (GCD)~\cite{gcd} is an emerging and challenging open-world problem that has received growing attention in recent years. The goal of GCD is to assign category labels to unlabeled samples that may belong to either known or unknown categories, by leveraging the knowledge contained in labeled data from known categories. Most existing GCD approaches focus on static images and have shown promising performance. However, these methods can be insufficient for tasks that inherently involve temporal dynamics. In particular, visually similar instances in individual frames may correspond to semantically different categories once temporal context is taken into account. For instance, as shown in~\cref{fig:1}, human actions such as ``Throwing firsbee'' and ``Catching firsbee'' can appear visually similar in isolated frames, yet are clearly distinguishable when their motion trajectories are considered.

To address this limitation, we extend the category discovery task to the video domain and introduce a new setting, termed \textbf{Video-GCD}, where both labeled and unlabeled samples are video clips enriched with temporal and motion cues. Compared to static image-based GCD, Video-GCD poses unique challenges: human/animal actions are intrinsically spatiotemporal, demanding models to jointly reason over appearance and dynamic motion patterns across frames. The absence of labeled data for unknown categories further compounds the difficulty, as learning discriminative temporal features becomes more ambiguous. These factors make Video-GCD a substantially more complex task, necessitating novel frameworks that can effectively capture and utilize temporal context for open-world action discovery.

To meet the requirement, we propose a novel framework called Memory-guided Consistency-aware Contrastive Learning (\textbf{MCCL}), which explicitly captures spatiotemporal cues and integrates them into contrastive learning through a consistency-guided voting mechanism. MCCL comprises two key components: Consistency-Aware Contrastive Learning (\textbf{CACL}) and Memory-Guided Representation Enhancement (\textbf{MGRE}). CACL models spatiotemporal differences as noise terms and realize the competition of time and space at the token level by learnable parameter between temporal-based and spatial-based tasks. By integrating multi-perspective spatiotemporal consistency voting, the approach first estimates consistency scores among unlabeled instances and optimizes inter-class boundaries. MGRE has built a dual layer memory buffer to store feature level and logical level representations that provide complementary information,  MGRE incorporates a dual-level memory buffer that stores both feature-level and logit-level representations that provide complementary information, alleviating the more severe problem of inter class confusion compared to image features. Moreover, to enable comprehensive evaluation, we introduce a new and challenging Video-GCD benchmark composed of 3 action recognition and 2 fine-grained bird classification datasets. 

Our main contributions are summarized as:
\begin{itemize}
    \item We propose a new GCD task that focuses on discovering categories in video, and accordingly builds a video-GCD benchmark including human actions and birds.
    
    \item We propose a MCCL framework as a strong baseline for video-GCD task, which can effectively integrate spatial and temporal cues. It is committed to alleviating the more severe inter class confusion caused by the entanglement of complex spatiotemporal features, and maximizing the construction of differential spatiotemporal features while ensuring the recognition accuracy of spatiotemporal collaboration, in order to optimize the clear decision boundaries between different classes in videos.
    
    \item Extensive experiments show that our proposed MCCL outperforms the advanced image-based GCD methods on both human action discovery and video-based bird discovery.
    
\end{itemize}

\section{Related Work}

\par\noindent
\textbf{Category Discovery Setups}. Category discovery aims to identify unknown categories from unlabeled data by transferring knowledge from a set of labeled categories. This task was initially studied as Novel Category Discovery (NCD)\cite{ncd}, which assumes that all unlabeled samples belong to unseen categories disjoint from the labeled ones. To address more realistic scenarios, Generalized Category Discovery (GCD)\cite{gcd} extends NCD by considering unlabeled data containing both known and unknown categories, enabling broader applicability in open-world settings. Beyond these category discovery settings, research has further explored extensions in diverse scenarios such as federated\cite{fgcd}, continual\cite{ma2024happy,cendra2024promptccd,zhang2022grow}, multi-modal\cite{clipgcd, textgcd, mgcd, get}, ultra-fine\cite{ultrafine}, and domain-shift-aware category discovery~\cite{wang2024hilo}, reflecting the growing interest in tackling category discovery under practical applications. \textit{In this work, we focus on addressing the GCD task in video, which needs to integrate both spatial and temporal cues to support more robust and meaningful class discovery.}

\textbf{Category Discovery Approaches}. Many methods have been proposed to solve the NCD tasks, such as~\cite{ncd, ncl, uno, dualrs, openmix, incd, ncdss, rankstats}. Specifically, early work~\cite{rankstats,dualrs} modeled the binary relationship between samples based on the similarity of the top-k feature dimensions. UNO~\cite{uno} built a connection of images and classes by the Sinkhorn-Knopp algorithm~\cite{sinkhorn}. Meanwhile, a line of work~\cite{ncl,openmix} employed richer inter-sample relations with nearest neighbor in feature space.  On the other hand, existing GCD approaches can be broadly categorized into parametric methods~\cite{simgcd, ugcd, sptnet, legogcd} and non-parametric methods~\cite{gcd, dccl, promptcal, gpc, pim, cms}. The former exhibits stable training, while the latter shows balanced representation ability between known and unknown categories.

\noindent\textbf{Semi-Supervised Action Recognition.} The exploration of self-supervised learning (SSL) in video recognition lags behind the progress in image classification. Due to the unique nature of the time dimension, positive and enhanced samples in contrastive learning need to adhere to temporal consistency while creating self-supervised signals throughout the time dimension. VideoSSL~\cite{jing2021videossl} compares SSL methods that are specifically applied to videos, revealing limitations in extending pseudo-labeling directly. LTG~\cite{LTG2022} utilizes the gradient mode of time to generate high-quality pseudo-labels for training. To address the limited temporal awareness in pixel-level augmentation methods like Mixup~\cite{mixup} and CutMix~\cite{cutmix}, SVFormer~\cite{xing2023svformer} captures token-level temporal correlations by maintaining a coherent masked token across the timeline. SeFAR~\cite{huang2025sefar} introduces a weak-to-strong consistency regularization mechanism within a teacher-student framework, enabling dual-level temporal element modeling for fine-grained action recognition. TimeBalance~\cite{dave2023timebalance}  leverages the temporal contrastive losses from TCLR~\cite{dave2022tclr} to learn the temporal distinctive teacher. \textit{Differ from these works that assume that the labeled and unlabeled videos share the same category space, our Video-GCD practically assumes that unlabeled videos may include previously unseen categories. Traditional Semi-Supervised action recognition leverages pseudo labels to supervise unlabeled data. Each category retains at least a few labeled samples, and under the assumption that all categories are known, this enables category representation learning and prototype construction. However, Video-GCD is oriented towards unknown categories, and due to the lack of referenceable label samples and category names, it can only determine the phylogenetic relationships between all samples and infer category structures through unsupervised/semi supervised generation of pseudo labels. The focus is on fundamentally discovering new category structures within clustering clusters, rather than traditional recognition tasks.}

\section{Methods}
\label{sec:methods}
\par\noindent
\textbf{Problem Setup.} Video-GCD aims to learn a model capable of accurately classifying unlabeled samples from known categories while simultaneously clustering those from unknown categories. Specifically, we consider a labeled video dataset $D_L = \{ \left( V_i^L, y_i^L \right) \}_{i=1}^{N_L}$, where $V_i^L$ denotes a labeled video consisting of $T$ frames, \textit{i.e.}, $V_i^L \in \mathbb{R}^{H \times W \times 3 \times T}$. Here, $y_i^L$ is the corresponding category label, and $N_L$ is the number of labeled videos. Similarly, we denote the unlabeled video dataset as $D_U = \{ \left( V_i^U, y_i^U \right) \}_{i=1}^{N_U}$, where $y_i^U$ is the ground-truth label for the $i$-th unlabeled video but is not available during training. Let $C_N$ denote the total set of categories in the entire dataset. We define $C_L$ as the set of categories present in the labeled dataset, and $C_U$ as the set of novel categories that do not appear in $C_L$, such that $C_N = C_L \cup C_U$ and $C_L \cap C_U = \emptyset$. Accordingly, the label space satisfies $y_i^L \in C_L$, while $y_i^U \in C_N = \{ C_L \cup C_U \}$. Following prior works~\cite{simgcd,rs}, we assume that the total number of categories $|C_N|$ is known in advance.

\par
\begin{figure*}[!t]
\centering{
\includegraphics[width=12cm]{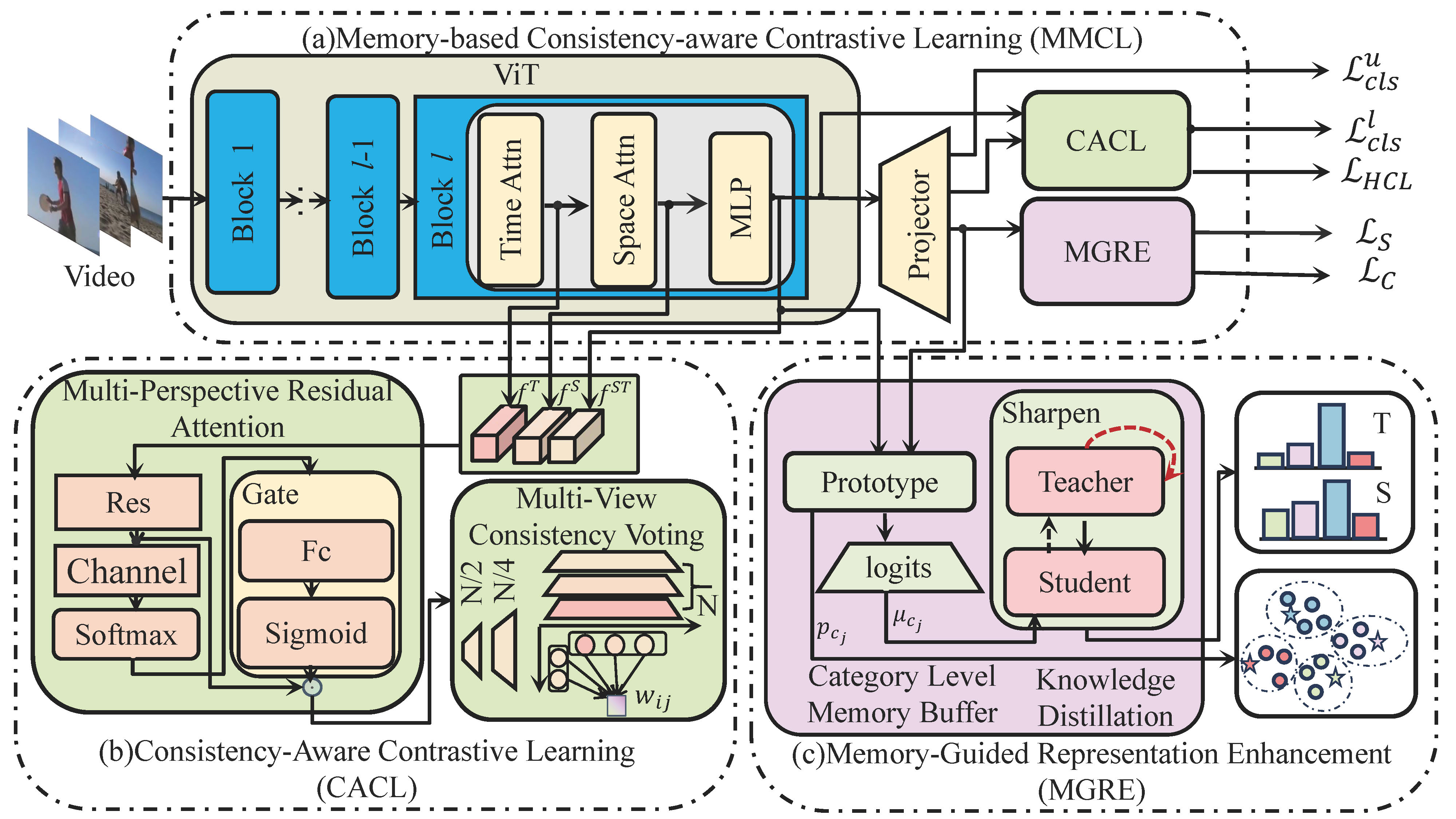}
}
\vspace{-1em}
\caption{Framework Overview of our Memory-based Consistency-aware Contrastive Learning (MCCL) framework. MCCL mainly consists of Consistency-Aware Contrastive Learning (CACL) and Memory-Guided Representation Enhancement (MGRE) module. CACL integrates spatial, temporal, and spatiotemporal information to measure consistency relationships between instances in a mini-batch. Then, CACL adopts a weighted contrastive loss based on the degree of consistency relationships. In Multi-View Consistency Voting, "N", "N/2", and "N/4" respectively represent the total number of categories in the corresponding voting layer. MGRE builds a dual-level memory buffer to store class-specific feature prototypes and logits prototypes. Then, MGRE encourages the model to generate compact clusters from both feature- and logits-level supervision, thereby enhancing the discriminative ability of representations. We elaborate on these modules in the following sections.\label{fig:2}}
\vspace{-1em}
\end{figure*}

\noindent
\textbf{Framework Overview.} As shown in~\cref{fig:2}, our Memory-guided Consistency-aware Contrastive Learning (MCCL) framework mainly consists of Consistency-Aware Contrastive Learning (CACL) and Memory-Guided Representation Enhancement (MGRE) module. CACL integrates spatial, temporal, and spatiotemporal information to measure consistency relationships between instances in a mini-batch. Then, CACL adopts a weighted contrastive loss based on the degree of consistency relationships, which effectively exploits the potential knowledge contained in both the labeled and unlabeled data, thereby improving the model's GCD ability. MGRE builds a dual-level memory buffer to store class-specific feature prototypes and logits prototypes. Then, MGRE encourages the model to generate compact clusters from both feature- and logits-level supervision, thereby enhancing the discriminative ability of representations. We elaborate on these modules in the following sections.

\subsection{Consistency-Aware Contrastive Learning (CACL)}
Although existing image-based GCD methods~\cite{simgcd,gcd} have shown that contrastive learning is effective for enhancing representation quality and improving GCD performance, we empirically observe that directly applying contrastive learning on spatiotemporal features from the backbone leads to suboptimal results, as evidenced in~\cref{tab:fusion}. This highlights a key challenge in Video-GCD: \textit{how to effectively incorporate multi-perspective spatiotemporal information into the contrastive learning.}
\par
To address this challenge, we first introduce a multi-perspective residual attention module that enhances the interaction among spatial, temporal, and spatiotemporal features. Based on this, we further propose a multi-perspective voting strategy to assess the consistency scores of instance pairs and inject such multi-view information into contrastive representation learning.

\par\noindent
\textbf{Multi-Perspective Residual Attention}
We propose a category-specific consistency-aware contrastive learning framework, which constructs a dynamic competition network between temporal and spatial tokens, achieves adaptive adjustment of spatiotemporal preference weights at the category level, and determines the competition proportion between spatial-based and temporal-based tasks.
Given a video sample $V_i\in D$, we separately extract temporal feature ${f}^{T}$, spatial feature ${f}^{S}$, and spatiotemporal feature ${f}^{ST}$ representations from the output of the final block in backbone network $E$~\cite{bertasius2021space}, which is formulated as: 
\begin{equation}
    f_{i}^{S}, f_{i}^{T}, f_{i}^{ST}=E(V_{i}).
\end{equation}

We define the competition between spatial and temporal as an additional learnable residual competition $f_i^{res}$ and capture their competitive interactions at each token position.To estimate the discriminative strength of each perspective channel, we define the residual ${f^{res}}$ between spatiotemporal features and a single feature as:
\begin{equation}
{f_{i}^{res}} = ({f}_{i}^{S} +{f}_{i}^{T}) - 2 \cdot {f}_{i}^{ST}
\end{equation}
We then compute the attention weights for each channel using Softmax normalization.
\begin{equation}
{w_i^{ST}} = \text{Softmax} \left( \frac{f_i^{ST}}{\tau \left( \sum_{c=1}^{C} {f}_{c}^{ST} + \varepsilon \right)} \right) \in \mathbb{R}^{C}, \quad \varepsilon=10^{-6}
\end{equation}
 The final residual-attentive fusion feature is obtained by linearly fusing the spatial and temporal features based on the learned attention weights:
\begin{equation}
f_{i}^{STF} = {f}_{i}^{ST} +\text{Gate}({f}_{i}^{ST}) \cdot{f_{i}^{res}} \cdot {w_{i}^{ST}},
\end{equation}
where the Gate$(\cdot)$ denotes the self-gate operation, which is implemented by a fully-connected layer followed by a Sigmoid function.
\par\noindent
\textbf{Discussion.} The $w_{i}^{ST}$ focuses on channels with drastic changes between three types of features, relying on the original data attributes to obtain spatiotemporal residuals. Channels with large numerical differences contain rich spatiotemporal information; The Gate$(\cdot)$ suppresses redundant channels and corrects them in a learnable manner, which may provide unreliable or redundant supervisory signals despite drastic changes.
\par\noindent
\textbf{Multi-View Consistency Voting.}
Sharing a similar motivation with image-based GCD methods~\cite{simgcd,gcd}, we adopt the InfoNCE~\cite{oord2018representation} to enhance representation learning in a semi-supervised manner. However, when applying it to unlabeled data, directly pulling an instance closer to its augmented counterpart while pushing it away from other instances often yields suboptimal performance on novel categories. This is primarily due to the absence of reliable supervisory signals for unlabeled samples. Thus, we propose to assign adaptive weights to instance pairs in contrastive learning based on their consistency scores, thereby providing more precise supervision for unlabeled data.

\par
Specifically, we first identify potentially consistent relationships between labeled and unlabeled instances by leveraging multiple clustering results. We consider two complementary groups of clusterings: 1) horizontal clusterings $\mathcal{C}^{hor}$ based on multi-perspective spatiotemporal features, and 2) vertical clusterings $\mathcal{C}^{ver}$ generated with varying numbers of clusters. The first group captures relational cues by analyzing discrepancies across spatial, temporal, and spatiotemporal features. The second group explores instance similarities at different levels of semantic granularity. To unify the processing of these results, we formulate a $k$th-level clustering assignment for the $i$-th instance as:

\begin{equation}
\mathbf{c}_i^k =
\begin{cases}
\mathcal{C}^{hor}\left( {f}_i^{*} , n\right), \text{where~} * \in \{S,T, STF\}, & \text{if } k\in \{0,1,2\} ,\quad n = |C_{N}| \\
\mathcal{C}^{ver}\left(f_i^{STF}, n\right), & \text{if } k \geq 3,\quad n = \frac{|C_{N}|}{2^{k-2}}.
\end{cases}
\end{equation}
\par
Building upon the clustering assignment, we propose to calculate a consistency score for each instance pair via a voting mechanism. Specifically, we define that the instance pairs that share the same clustering assignment across more levels have a higher consistency. Here, we consider voting across $K$ levels, and the normalized  consistency score $c_{ij}$ is formulated as:

\begin{equation}
c_{ij} = (1 - \eta) \cdot y_{ij} + \eta \cdot \frac{w_{ij}}{\sum\limits_{k \ne i} w_{ik}},      w_{ij} = \sum_{v=1}^{K} {\mathbbm{1}(c_i^v = c_j^v)}, y_{ij} = \begin{cases}
1, & \text{if } i~\text{and}~j \in {D}_{L}, \\
0, & \text{either } i ~\text{or}~ j \in {D}_{U},
\end{cases}
\label{eq:6} 
\end{equation}
where $\mathbbm{1}$ is an indicator function with a value of 1 if it is matched and 0 if it is not. The $w_{ij}$ is the voting statistics between sample $i$ and $j$, where $y_{ij}$ is the binary label indicating whether $j$ is a positive sample of $i$, $ \eta$ is the trade-off factor that control the learning strength of labeled and unlabeled data.

\par\noindent

\textbf{Consistency-Aware Contrastive Loss}.
Let $\mathcal{B^I}$ be the set of all other samples except $V_{i^{\prime}}$. The consistency-aware contrastive loss is defined as:
\begin{equation}
\mathcal{L}_{HCL} = - \sum_{{j^{\prime}} \in \mathcal{B}^I, {j^{\prime}} \ne {i^{\prime}}} 
 c_{{i^{\prime}}{j^{\prime}}} \cdot 
\log \left(
\frac{
\exp \left( \frac{ f_{i^{\prime}}^\top f_{j^{\prime}} }{ \tau_H } \right)
}{
\sum\limits_{k \in \mathcal{B}^I, k \ne {i^{\prime}}} 
\exp \left( \frac{ f_{i^{\prime}}^\top f_k }{ \tau_{H_I} } \right)
}
\right),
\label{eq:7} 
\end{equation}
where $\tau_H$,$\tau_{H_I}$ are temperature parameters controlling the sharpness of the distribution. This formulation emphasizes hard negatives and confident positives through the learned weight $w_{{i^{\prime}}{j^{\prime}}}$. It should be noted that $c_{ij}$ in Eq.~\ref{eq:6} is the semantic similarity determined through multi-view voting among all samples to dynamically adjust the negative sample weights of each sample; And $c_{i^{\prime}j^{\prime}}$ in Eq.~\ref{eq:7} represents the subset of $c_{ij}$ based on the inter batch sample relationship.

\subsection{Memory-Guided Representation Enhancement (MGRE)}
Due to the complex spatiotemporal characteristics of videos, features extracted from instances of the same category often exhibit substantial variations. To address this issue, we propose to enhance feature representation via dual-level distillation from a category-level memory buffer.

Prototypes, typically computed as the means of class-wise features, provide strong supervision by capturing intra-class commonality through cross-entropy. However, direct averaging may introduce noise and increase inter-class confusion, while lacking awareness of the overall inter-class structure\cite{LDClogitsdeConfusion}. In contrast, logits in Transformers aggregate global context, and their attention maps highlight key visual cues for decision-making~\cite{yang2025visionzip,vasu2025fastvlm}. Therefore, we propose a complementary bipolar supervision strategy: using distilled logits to guide prototype learn logit deconfusion, and enhance inter-class boundary discrimination.

\par\noindent
\textbf{Category-Level Memory Buffer.} To fully utilize the knowledge of labeled videos, we build category-level memory to store both feature- and logits-level category prototypes. Specifically, we randomly sample a small portion of the labeled data set $D_L$ to form a representative subset $D_M =\left\{ \left( V_i, y_i^L\right) \right\}_{i=1}^{N_M}$ and $C_M\subseteq C_L$. For each known category ${c_j}\in{C_M}$, its unique category label is ${c_j}$. To capture the commonalities of the ${c_j}$ category, we compute the feature prototype $p_{c_j}$ by averaging the spatiotemporal features $f_i^{ST}$ whose labels $y_i= c_j$, which is formulated as:
\begin{equation}
p_{c_j} = \frac{1}{|\mathcal{I}_{c_j}|} \sum_{i\in \mathcal{I}_{c_j}}f_{i}^{ST} \quad, \quad \mathcal{I}_{c_j}=\{i\mid y_i = c_j\},
\end{equation}
where $\mathcal{I}_{c_j}$ denotes the index set of samples belonging to class $c_j$, and $|\mathcal{I}_{c_j}|$ is the number of samples in that category. 
\par\noindent
\textbf{Feature Prototype Knowledge Distillation.}
To distill comprehensive category knowledge from prototype to corresponding instances, we define a contrastive loss to align the instance features with the prototype in the representation space, which is formulated as:
\begin{equation}
{L}_C = -\log \frac{\exp\left(f \cdot p_{{c}_j} / \tau_{C_L} \right)}{\sum_{k=1, k \neq j}^{|C_M|} \exp\left({f} \cdot p_{ck} / \tau_{C_L} \right)},
\end{equation}

where $\tau_{C_L}$ is the temperature coefficient. $|C_M|$ is the number of categories in $D_M$.

\par\noindent
\textbf{Logits Prototype Knowledge Distillation.}
Apart from the information encoded in feature prototypes, in this work, we consider additional information contained in logits prototypes. 
\par
First, we feed the feature prototypes to classifier $G$ to generate logits prototypes. Meanwhile, we pass $f_i^{STF}$ feature to $G$ to yield instance logits $z_{i}$. Formally, this process is defined as:
\begin{equation}
\mu_{c_j} = G(p_{c_j}),   z_{i} = G(f_i^{STF})
\end{equation}


\par
Second, we employ Kullback-Leibler (KL) divergence to transfer knowledge to the instance logits $z_i $ from the stored concept logits $\mu_{c_i} $. To encourage the model's predictions to align more closely with the target class's average prediction distribution, we sharpened the category logic of the teacher by utilizing the temperature $\tau_{T_L} < 1$:

\begin{equation}
\tilde{{\mu}}_{c_j} = \mathrm{Softmax} \left( \frac{{\mu}_{c_j}}{\tau_{T_L}} \right),
\end{equation}

Third, we design the prototype-based logit distillation loss to enhance category-level supervision at the logits level, which is defined as:

\begin{equation}
\mathcal{L}_S = \frac{1}{B} \sum_{i=1}^B  {\tau_{S_L}} \cdot \mathrm{KL} \left( \mathrm{LogSoftmax} \left({z}_i \right) \;\|\; \tilde{{\mu}}_{c_j} \right), y_i = c_j,
\end{equation}
where the $\tau_{S_L}$, $\tau_{T_L}$ is the distillation temperature to smooth logits. The $\mathrm{KL}(\cdot \|\cdot)$ denotes the Kullback-Leibler divergence. The $B$ is the batch size.

\subsection{Optimization and Inference}
To facilitate clustering-friendly representation learning, we employ $\mathcal{L}_{\text{cls}}^{s}$  and $\mathcal{L}_{\text{cls}}^{u}$ in the advanced SimGCD~\cite{simgcd} to exploit both labeled and unlabeled data. The total loss is formulated as follows:

\begin{equation}
\mathcal{L} = \lambda_{Sup} \cdot (\mathcal{L}_{\text{cls}}^{s}+\mathcal{L}_C+ \lambda_S \cdot \mathcal{L}_S) + \lambda_{Unsup} \cdot (\mathcal{L}_{\text{cls}}^{u}+\mathcal{L}_{HCL}),
\end{equation}
where the $\lambda_{Sup}$, $\lambda_{UnSup}$ are the balance weight between supervised and unsupervised learning. The $\lambda_{S}$ is the weight factor for controlling the strength of $ \mathcal{L}_S$.

\textbf{Inference.} During testing, we extract all spatiotemporal features $f^{ST}$ from the backbone network, following ~\cite{gcd}. We then perform K-Means clustering to assign cluster indices to all instances, and apply the Hungarian matching algorithm to compute the accuracy between the predicted cluster indices and the ground truth labels. The \textit{All ACC}, \textit{Old ACC}, and \textit{New ACC} are reported.

\section{Experiments}
\subsection{Experiment Setup}
\par\noindent

\textbf{Video-GCD Benchmark.} As shown in~\cref{tab:dataset}, we construct a new benchmark for the Video Generalized Category Discovery (Video-GCD) task by reorganizing five video datasets: three general human activity datasets (\textit{e.g.,} UCF101~\cite{ucf101}, SSv2~\cite{goyal2017ssv2}, and Kinetics-400~\cite{carreira2017k400} and two fine-grained bird datasets (\textit{e.g.,} VB100~\cite{vb100} and IBC127~\cite{ibc127}). For category split protocols, we adopt the following strategy: except for VB100~\cite{vb100}, where the first 50\% of classes are used as known and the remaining 50\% as unknown, we designate even-indexed classes as known and odd-indexed classes as unknown in all other datasets means "even-odd". In terms of data volume, we utilize the full dataset for UCF101~\cite{ucf101}, VB100~\cite{vb100}, and IBC127~\cite{ibc127}, while sampling 15\% of the total videos from SSv2~\cite{goyal2017ssv2} and Kinetics-400~ \cite{carreira2017k400} to reduce computational cost while retaining diversity. Specifically, \textbf{UCF101}~\cite{ucf101} consists of 101 human action categories (e.g., makeup, diving, horse riding), with a total of 9,935 videos and 5–22 clips per category.
\textbf{SSv2}~\cite{goyal2017ssv2} focuses on fine-grained object manipulation tasks (e.g., “pushing something out,” “taking something away”), emphasizing temporal reasoning and motion perception.
\textbf{Kinetics-400}~\cite{carreira2017k400} features 400 action categories from YouTube videos, such as “playing football” and “playing the piano.” Although large-scale, many actions are recognizable from static frames.
\textbf{VB100}~\cite{vb100} includes 1,416 clips across 100 bird species. Each species has an average of 14 clips, with 798 captured using handheld devices and the remainder from static cameras.
\textbf{IBC127}~\cite{ibc127} comprises 8,014 clips spanning 127 fine-grained bird categories. Each category contains between 21 and 226 clips, showcasing common bird behaviors like foraging, grooming, and feeding. \textit{This diverse benchmark enables a comprehensive evaluation of Video-GCD methods across both general and fine-grained action recognition tasks in open-world settings.}

\begin{table*}[t]
\renewcommand\arraystretch{1.1}
\setlength\tabcolsep{1.3pt}
\centering
\caption{Detailed Statistics on 5 video-GCD datasets. ``UnK-C.'' and ``K-C.'' denote ``unknown-category'' and ``known-category'', respectively.\label{tab:dataset}}
\begin{tabular}{c|c|c|c|c}
\hline
\hline
    \multicolumn{1}{c|}{Dataset} & \multicolumn{1}{c|}{\# All Video} & \multicolumn{1}{c|}{\# K-C. Video} & \multicolumn{1}{c|}{\# UnK-C. Video} & \multicolumn{1}{c}{Category Division}\\   
\hline
UCF101\cite{ucf101} & 13190 &2908  &9935 &100,(even,odd) \\
SSV2\cite{goyal2017ssv2}   &25253  &5528  &19086 &174,(even,odd)\\ 
Kinetics400\cite{carreira2017k400}   &34828  &29516 &7974 &400,(even,odd)\\ 
VB100\cite{vb100}   &7280  &1556  &5539 &(0,50),(51,100) \\ 
IBC127\cite{ibc127}   &7616  &1840 &5592 &127,(even,odd)\\ 
\hline 
\end{tabular}%
\vspace{-1.5em}
\end{table*}%

\par\noindent
\textbf{Training Details}. Our training is divided into two stages. In the first stage of training, ImageNet1k~\cite{imagenet} is used as a pre-trained weight to initialize the ViT~\cite{arnab2021vivit} backbone network with a depth of 8; we referred to the settings of TimeSformer~\cite{2021timesformer} and only use labeled data for cross-entropy training; Our initial learning rate is 0.005, momentum is set to 0.9, weight decay is set to 0.0001, and we use SGD optimizer. The size of the cropped area per frame fed into the network is 224 × 224, and we sample 8 frames per video sample, \textit{i.e.,} $T=8$. In the second stage, unlabeled data is added for training. Following SimGCD~\cite{simgcd}, our Projector $G$ adopts DINO~\cite{dino} structure with 3 layers of MLP and a feature input dimension of 768. Its output dimension is the number of categories. Aligned with~\cite{2024selex}, temperatures $\tau_H$ and $\tau_{H_I}$ are both set to 1.0,while temperatures $\tau_{C_L}$ and $\tau_{H_{T_L}}$ are set to 0.05 and 0.1. Except for K400 which uses 4 4090 GPUs, the other datasets use 2 4090 GPUs.


\par\noindent
\textbf{Evaluation Metrics.} We adopt \textit{All ACC}, \textit{Old ACC}, and \textit{New ACC} as our evaluation metrics. During testing, we first extract video features using the trained TimeSformer~\cite{2021timesformer} model. Then, K-means clustering is applied to group the features, and we compute clustering accuracy by aligning the predicted clusters with ground-truth labels via the Hungarian matching algorithm. \textbf{All ACC} measures the overall classification accuracy across all unlabeled samples, providing a holistic assessment of the model’s performance. \textbf{Old ACC} reports the classification accuracy on unlabeled samples that belong to known categories, reflecting the model’s ability to generalize to known classes. \textbf{New ACC} evaluates the classification accuracy on samples from novel categories, indicating the effectiveness of the method in discovering new categories. Together, these metrics offer a comprehensive evaluation of the model’s ability to handle both seen and unseen categories in an open-world setting.

\subsection{Comparison with Strong Baselines Adapted from Image-based GCD Methods}
Since Video-GCD is a new task, we adapt the advanced image-based GCD methods, \textit{e.g.}, SimGCD~\cite{simgcd}, InfoSieve~\cite{2023Sieve}, SPTNet~\cite{sptnet}, and SelfEx~\cite{2024selex} into our framework as competitive baseline methods. Straightforwardly, we replace the image feature used in these methods with the spatiotemporal features $f^{ST}$ output from the last block of TimeSformer~\cite{2021timesformer}. We report the detailed results in~\cref{tab:action} and~\cref{tab:bird}.

\begin{table*}[t]
\renewcommand\arraystretch{1.1}
\setlength\tabcolsep{1.3pt}
\centering
\caption{Comparison experiments on 3 action recognition datasets.}
\begin{tabular}{l|c|ccc|ccc|ccc|ccc}
\hline
\hline
     \multirow{2}{*}{Method} &\multirow{2}{*}{Venue}& \multicolumn{3}{c}{UCF101} & \multicolumn{3}{c}{SSV2} & \multicolumn{3}{c}{Kinetics400} & \multicolumn{3}{c}{Average}\\
    \cmidrule(lr){3-5} \cmidrule(lr){6-8} \cmidrule(lr){9-11} \cmidrule(lr){12-14} 
  &  &\multicolumn{1}{c}{All}   &\multicolumn{1}{c}{Old} &\multicolumn{1}{c}{ New} 
&\multicolumn{1}{c}{All} &\multicolumn{1}{c}{Old} &\multicolumn{1}{c}{New} &\multicolumn{1}{c}{All} &\multicolumn{1}{c}{Old} &\multicolumn{1}{c}{New} &\multicolumn{1}{c}{All} &\multicolumn{1}{c}{Old} &\multicolumn{1}{c}{New} \\
\hline    
SimGCD~\cite{simgcd}&ICCV'23 & 63.93  & 82.65 & 54.70 & 11.36 & 14.27 & 9.93 & 21.57 & 25.15 & 19.79 & 32.29 & 40.69 & 28.14 \\
InfoSieve~\cite{2023Sieve}&NIPS'23 &58.02 &69.90 & 52.60  & 11.10 &12.92 &10.22 &21.28 &22.28 &\textbf{20.78} &30.13 &35.70 &27.87 \\
SPTNet~\cite{sptnet}&ICLR'24 &38.50 &55.69 &30.03  &7.92 &10.11 &6.85 &12.55 &14.32 &11.67 &19.66 &26.71 &16.18 \\
SelfEx~\cite{2024selex}&ECCV'24 &63.43 &77.07 & 56.72  & 12.54 & 14.96 & 11.35 & 21.23 & 23.14 & 20.28 &32.40 &38.39 &29.45\\
MCCL  &Ours & \textbf{68.27} &\textbf{89.54} &\textbf{57.80} &\textbf{13.58} &\textbf{18.69} &\textbf{11.57} &\textbf{22.95} &\textbf{29.89} & 19.51 &\textbf{34.93} &\textbf{46.04} &\textbf{29.63} \\  \hline 
\end{tabular}%
\label{tab:action}%
\vspace{-1.5em}
\end{table*}%

\begin{table*}[t]
\renewcommand\arraystretch{1.1}
\setlength\tabcolsep{1.3pt}
\centering
\caption{Comparison experiments on 2 fine-grained classification video datasets.}
\begin{tabular}{l|c|ccc|ccc|ccc}
\hline
\hline
    \multirow{2}{*}{Method}&\multirow{2}{*}{Venue} & \multicolumn{3}{c}{VB100}  & \multicolumn{3}{c}{IBC127}& \multicolumn{3}{c}{Average}\\
    \cmidrule(lr){3-5} \cmidrule(lr){6-8} \cmidrule(lr){9-11} 
  &  &\multicolumn{1}{c}{All}   &\multicolumn{1}{c}{Old} &\multicolumn{1}{c}{ New} 
&\multicolumn{1}{c}{All} &\multicolumn{1}{c}{Old} &\multicolumn{1}{c}{New}&\multicolumn{1}{c}{All} &\multicolumn{1}{c}{Old} &\multicolumn{1}{c}{New} \\
\hline   
SimGCD~\cite{simgcd}&ICCV'23 & 33.75 & 37.66 & 32.01 & 31.88 & 31.31 & 32.16 & 32.18 & 34.18 & 31.33 \\
InfoSieve~\cite{2023Sieve}&NIPS'23 & 33.71 & 37.14 & 32.12 & 31.16 & 29.38 & 32.00 & 32.44 & 33.26 & 32.06 \\
SPTNet~\cite{sptnet}&ICLR'24 & 23.88 &25.83 &21.22 &19.91 &20.82 &19.45 &21.90 &23.33 & 20.78 \\
SelfEx~\cite{2024selex}&ECCV'24 & 40.43 & 47.83 & 33.68 & 33.59 & 33.76 & \textbf{33.50} & 37.01 & 40.80 & 33.59 \\
MCCL & Ours & \textbf{46.17} & \textbf{70.51} & 34.81 & \textbf{38.06} & \textbf{48.83} & 32.72 & \textbf{42.12} & \textbf{59.67} & \textbf{33.77} \\  \hline 
\end{tabular}%
\label{tab:bird}%
\vspace{-1.5em}
\end{table*}%
\par\noindent\textbf{Result Summary.} As shown in~\cref{tab:action} and~\cref{tab:bird}, we can clearly observe that on most datasets, our Video-GCD outperforms other models in all aspects. SPTNet~\cite{sptnet} performs best on SSv2~\cite{goyal2017ssv2}, our algorithm performs suboptimal in known category recognition, which means we need to alleviate forgetting of existing knowledge in future work. The All acc of Video-GCD on UCF101~\cite{ucf101}, Kinetics400~\cite{carreira2017k400}, VB100~\cite{vb100}, and IBC127~\cite{ibc127} reached 68.27$\%$, 22.95$\%$, 46.17$\%$, 38.06$\%$, respectively; Compared with other suboptimal high-precision algorithms, its accuracy exceeds 4.34$\%$,1.38$\%$, 5.74$\%$, and 4.47$\%$, respectively. 

\subsection{Ablation Study}
To verify the effectiveness of the proposed methods, we conduct two group experiments and report the results in~\cref{tab:label4} and~\cref{tab:label5}, respectively. The ``Baseline'' method debote we use only the $\mathcal{L}_{cls}^{s}$ and $\mathcal{L}_{cls}^{u}$ in ~\cite{simgcd} to train our model. The ablation experiment in~\cref{tab:label4} verifies the effects of $MGRE$ and $CACL$ in the proposed framework. When only employing the $MGRE$, all acc improves by 0.35$\%$, 1.41$\%$, 10.27$\%$, and 4.21$\%$ on datasets UCF101\cite{ucf101}, SSV2\cite{goyal2017ssv2}, VB100\cite{vb100}, and IBC127\cite{ibc127}, respectively. When only the $CACL$ model is applied, the improvement impacts are 1.49$\%$, 5.75$\%$, and 2.08$\%$, respectively. In Table~\ref{tab:label5}, we conducted experiments on VB100~\cite{vb100} for various segmented structures. It can be seen that each module contributes to the improvement of ALL ACC. Multi-View Consistency Voting.
($\mathcal{L}_{HCL}$) and Category Level Memory Buffer
 ($\mathcal{L}_{C}$) have significant effects in improving known categories, while Knowledge Distillation
($\mathcal{L}_{S}$) and Multi-Perspective Residual Attention (\text{MPRA}) contribute to new categories. An intriguing pattern emerges from our analysis: while $\mathcal{L}_{S}$ significantly enhances performance on old classes at the potential cost of new class accuracy, $\mathcal{L}_{C}$ provides complementary benefits by improving new class recognition. Their integration within MGRE enables joint performance gains across both class types. A similar synergy is observed within CACL, where $\mathcal{L}_{HCL}$  and MPRA collaboratively contribute to improved overall results through their distinct yet complementary effects.

\begin{table*}[t]
\renewcommand\arraystretch{1.3}
\setlength\tabcolsep{1.4pt}
\centering
\caption{Ablation study on our CACL and MGRE modules. \label{tab:fusion}}
\begin{tabular}{l|ccc|ccc|ccc|ccc}
\hline
\hline
    \multirow{2}{*}{Method} & \multicolumn{3}{c}{UCF101} & \multicolumn{3}{c}{SSV2}  & \multicolumn{3}{c}{VB100}  & \multicolumn{3}{c}{IBC127}\\
    \cmidrule(lr){2-4} \cmidrule(lr){5-7} \cmidrule(lr){8-10} \cmidrule(lr){11-13}
\multicolumn{1}{c}{}   &\multicolumn{1}{|c}{All}   &\multicolumn{1}{c}{Old} &\multicolumn{1}{c}{ New} 
&\multicolumn{1}{c}{All} &\multicolumn{1}{c}{Old} &\multicolumn{1}{c}{New} &\multicolumn{1}{c}{All} &\multicolumn{1}{c}{Old} &\multicolumn{1}{c}{New} &\multicolumn{1}{c}{All} &\multicolumn{1}{c}{old}
&\multicolumn{1}{c}{New} \\
\hline   
Baseline & 63.93  & 82.65 & 54.70 & 11.36 & 14.27 & 9.93  & 33.75 & 37.66 & 32.01 & 31.88 & 31.31 & 32.16 \\
+ MGRE & 64.28 &88.66 &52.27 &12.77 &16.83 &10.78  & 44.02 &67.57  &33.02 &36.09 & 43.23 & 32.54\\
+ CACL &65.42 &82.37 &57.06  & 11.31 &14.45 &9.76  &39.50 &50.62 &34.31 & 33.96 & 31.98 & \textbf{34.94}\\
Full Ours    & \textbf{68.27} &\textbf{89.54} & \textbf{57.80} & \textbf{13.58} &\textbf{18.69} & \textbf{11.57}  & \textbf{46.17} &\textbf{70.51} & \textbf{34.81} &\textbf{38.06} & \textbf{48.83} & 32.72 \\ 
\hline 

\end{tabular}%
\label{tab:label4}%
\vspace{-1.5em}
\end{table*}


\begin{table*}[t]
\renewcommand\arraystretch{1.3}
\setlength\tabcolsep{2pt}
\centering
\caption{\label{tab:ablation_deltas}Ablation study with step-by-step improvements on the VB100 dataset.}
\begin{tabular}{l l c c c c c c c}
\toprule[1.2pt]
Module & Configuration & 
\multicolumn{1}{c}{$\mathcal{L}_S$} & 
\multicolumn{1}{c}{$\mathcal{L}_C$} & 
\multicolumn{1}{c}{$\mathcal{L}_{HCL}$} & 
\multicolumn{1}{c}{MPRA} & 
All ACC & Old ACC & New ACC \\
\midrule
\cmidrule(lr){3-6}
Baseline &  & $\times$ & $\times$ & $\times$ & $\times$ & 33.75 (--) & 37.66 (--) & 32.01 (--) \\
\midrule
\multirow{2}{*}{MGRE} 
& \quad $+\mathcal{L}_S$ & $\surd$ & $\times$ & $\times$ & $\times$ & 43.75 \textbf{(+10.00)} & 70.75 \textbf{(+33.09)} & 31.05 \textbf{(-0.96)} \\
& \quad $+\mathcal{L}_C$ & $\surd$ & $\surd$ & $\times$ & $\times$ & 44.02 \textbf{(+0.27)} & 67.57 \textbf{(-3.18)} & 33.02 \textbf{(+1.97)} \\
\midrule
\multirow{2}{*}{CACL} 
& $+\mathcal{L}_{HCL}$ & $\surd$ & $\surd$ & $\surd$ & $\times$ & 44.32 \textbf{(+0.30)} & \textbf{73.45} \textbf{(+5.88)} & 30.71 \textbf{(-2.31)} \\
& +MPRA & $\surd$ & $\surd$ & $\surd$ & $\surd$ & \textbf{46.17} \textbf{(+1.85)} & 70.51 \textbf{(-2.94)} & \textbf{34.81} \textbf{(+4.10)} \\
\bottomrule[1.2pt]
\end{tabular}
\label{tab:label5}%
\vspace{-1em}
\end{table*}

\begin{figure*}[!t]
\centering{
\includegraphics[width=14cm]{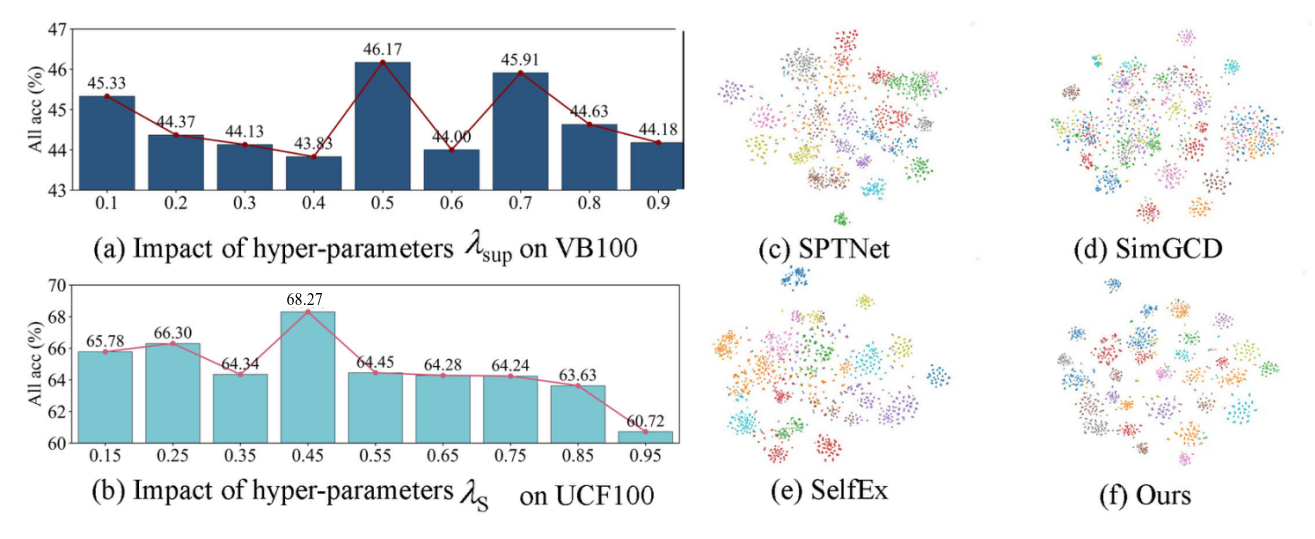}
}
\vspace{0.5em}
\caption{\label{fig:3}(a) and (b) illustrate the impact of hyper-parameters $\lambda_{s}$ and $\lambda_{Sup}$. (c-f) is the visualization of clustering effects with comparison methods.}
\end{figure*}

\subsection{Further Exploration }

\par\noindent\textbf{Hyper-Parameter Analyses}. We discuss the impact of hyperparameters in our MCCL, including the proportion of supervised weight loss $\lambda_{Sup}$ and loss weight $\lambda_s$. We select the optimal value of $\lambda_s$, and the impact of different values is depicted in Fig.~\ref{fig:3}(a). It achieved the highest 46.17$\%$ when $\lambda_{s} = 0.5$. Subsequently, we tend to utilize larger $\lambda_{Sup}$ to provide more reliable information. The impact of different values of $\lambda_{Sup}$ is shown in Fig.~\ref{fig:3}(b). It can be seen that the highest accuracy is achieved when $\lambda_{Sup} = 0.45$.  
\par\noindent\textbf{Visualization.} \cref{fig:3} (c-f) shows the clustering visualization effect of our algorithm compared to other advanced algorithms on the UCF101 dataset. It can be seen that our clustering is more compact internally, while the distribution of different clustering centers maintains a longer distance.

In real-world scenarios, the number of categories (i.e., $K$) is often unknown, especially for novel categories. The assumption about using ground-truth $K$ to train a fixed classifier is not practical. Thus, we follow the strategy in~\cite{vaze2022generalized} to estimate the  $K$ before second-stage training and set the $K$ to the estimated $\tilde{K}$ for training and testing.

We present the training results of the baseline algorithm using the estimated $\tilde{K}$ values on four datasets in ~\cref{tab:table11}. During testing, we calculated the clustering performance based on these estimated $\tilde{K}$ values. Except for VB100~\cite{vb100}, the predicted number of categories $\tilde{K}$ is generally smaller than the ground-truth. On the IBC127~\cite{ibc127} and SSv2~\cite{goyal2017ssv2} datasets, there is a considerable discrepancy between the predicted and actual number of categories (IBC127~\cite{ibc127}: 127 ground-truth classes vs. 92 predicted; SSv2~\cite{goyal2017ssv2}: 174 ground-truth vs. 134 predicted). This deviation may stem from high semantic similarity between fine-grained categories,  which poses challenges given the limited discriminative capacity of video features. Future work should further enhance the accuracy of category estimation.

\begin{table}[htbp]
\renewcommand\arraystretch{1.1}
\setlength\tabcolsep{1.3pt}
\centering
\caption{K-Unknown Performance and Estimation of the Number of Categories on 4 Datasets by the Estimation Method in~\cite{vaze2022generalized}.}
\begin{tabular}{l|c|ccc|ccc|ccc|ccc}
\hline
\hline
     \multirow{2}{*}{Method} &\multirow{2}{*}{Input}& \multicolumn{3}{c}{UCF101}& \multicolumn{3}{c}{SSv2} & \multicolumn{3}{c}{VB100}& \multicolumn{3}{c}{IBC127}\\
     & & \multicolumn{3}{c|}{$\tilde{K}=84$} & \multicolumn{3}{c|}{$\tilde{K}=134$} 
     & \multicolumn{3}{c|}{$\tilde{K}=103$} & \multicolumn{3}{c}{$\tilde{K}=92$} \\
    \cmidrule(lr){3-5} \cmidrule(lr){6-8} \cmidrule(lr){9-11} \cmidrule(lr){12-14} 
  &  &\multicolumn{1}{c}{All}   &\multicolumn{1}{c}{Old} &\multicolumn{1}{c}{ New} 
&\multicolumn{1}{c}{All} &\multicolumn{1}{c}{Old} &\multicolumn{1}{c}{New} &\multicolumn{1}{c}{All} &\multicolumn{1}{c}{Old} &\multicolumn{1}{c}{New} &\multicolumn{1}{c}{All} &\multicolumn{1}{c}{Old} &\multicolumn{1}{c}{New} \\
\hline   
Baseline &$B_{feat}$ &\textbf{61.35} &65.14 &\textbf{59.48} & 11.84 & 11.94 & \textbf{11.79} & \textbf{44.49} &\textbf{64.39} &\textbf{35.80}  &\textbf{33.27} & 29.74 &\textbf{35.00}  \\
Baseline &$Logit$  &50.75 &\textbf{83.75} &34.50 & \textbf{14.57} &\textbf{21.28} &11.27 & 30.70 &29.78 &31.13  &30.61 &\textbf{47.61} &22.18\\
\hline 
\end{tabular}%
\label{tab:table11}%
\end{table}%

\section{Conclusion}
We introduce Video-GCD, an extension of Generalized Category Discovery (GCD) to the video domain, addressing the limitations of static image-based methods in capturing temporal dynamics. To tackle the added complexity of spatiotemporal patterns, we propose MCCL, a novel framework combining Consistency-Aware Contrastive Learning (CACL) and Memory-Guided Representation Enhancement (MGRE). MCCL estimates instance consistency via multi-perspective temporal features and enhances representations with dual-level memory. We build a new benchmark covering both action and bird behavior datasets, which can serve as a starting point for future Video-GCD research.


{
    \bibliographystyle{unsrt}
\bibliography{gcd}
}





\end{document}